# Artificial intelligence and biosecurity: capabilities, threat pathways, and defense-in-depth governance

Candace S.Y. Chan[1], Aris Karatzikos[1], Ilias Georgakopoulos-Soares[1,*]

[1] Division of Pharmacology and Toxicology, College of Pharmacy, The University of Texas at Austin, Dell Paediatric Research Institute, Austin, TX, USA.

*correspondence to: ilias@austin.utexas.edu

## Abstract

Artificial intelligence is reshaping biological research across an increasingly connected digital-to-physical workflow. General-purpose large language models can retrieve and integrate scientific information, support experimental planning, and computational analysis; biological foundation models can predict, optimize, and generate proteins, genes, and genome-scale sequences; agentic systems can coordinate multistep research tasks; automated laboratories can partially close the design-build-test-learn cycle. These technologies could greatly benefit medicine, public health, and biotechnology. However, their biosecurity risk depends not only on what the AI can do, but also on who uses it, their expertise and intent, their access to laboratory tools and materials, and the safeguards in place. Current evidence shows that AI uplift exists but primarily affects digital rather than physical tasks. Frontier systems have exceeded expert baselines on *in-silico*, and screening-evasion benchmarks, whereas controlled wet-laboratory studies find that tacit knowledge and physical execution remain substantial barriers. This review describes the different biological threats from AI tool use, from information gathering and biological design to procurement, synthesis, testing, scale-up, and potential release. We further examine why alignment techniques for general-purpose models transfer poorly to biological ones, and the emerging role of interpretability in auditing whether hazardous capabilities are genuinely removed. We argue for defense-in-depth governance that links capability thresholds to proportionate responsibilities across the biological AI ecosystem, reducing high-consequence risk while preserving beneficial use.

**Keywords:** biosecurity; artificial intelligence; biological foundation models; large language models; genomic language models; protein design; laboratory automation; nucleic acid synthesis; dual-use research; governance

## 1. Introduction

Recent advances in biological AI, including large language models (LLMs), genomic language models (gLMs), protein language models (pLMs), agentic systems, and AI-enabled autonomous laboratory environments, can accelerate discovery, improve the design of biological systems, and transform experimental workflows. These technologies encompass several distinct but increasingly interconnected classes of systems. General-purpose LLMs can retrieve and integrate scientific information, support experimental planning, generate code, and assist with troubleshooting [1,2]. Biological foundation models trained on proteins or genomic sequences can predict molecular properties, evaluate mutations, and generate proteins, genes, or longer genomic sequences [3–6]. Agentic systems extend these capabilities by decomposing research objectives into multistep workflows and interacting with external computational tools [7]. Laboratory automation and nucleic-acid synthesis accelerate mechanisms through which digital designs can be experimentally implemented [8,9].

For example, during the SARS-CoV-2 pandemic, protein language models were used across multiple areas of virology and biomedical research to predict viral fitness and immune escape and to accelerate vaccine and treatment development [10]. These applications demonstrate the potential of biological AI to improve public health. However, the same capabilities that enable researchers to analyze and engineer biological systems could also be misused, potentially creating biosecurity risks.

These developments are particularly important because AI developers and policymakers have not established clear and consistently applied boundaries around which model capabilities create the greatest biosafety or biosecurity risks. Without such boundaries, it is difficult to anticipate the capabilities of new or combined AI systems and determine which risks should be prioritized for mitigation [11]. An analysis identified 1,107 AI-enabled biological tools across 76 countries and found that almost as many tools were released during 2023-2024 as in the previous four years combined. Among 57 state-of-the-art tools selected for detailed assessment, 82.5% had at least one open-source component, while 61.5% of tools classified as potentially dangerous were fully open-source. The analysis found no correlation between misuse-relevant capabilities and tool accessibility [12]. International committees seeking to establish guidelines against biotechnology misuse have also faced challenges, including inadequate funding, weak enforcement strategies, and the rapid pace of advances in AI [13].

Developers have attempted to safeguard AI models against misuse through approaches such as excluding sensitive data from pretraining datasets, selectively unlearning hazardous or virus-related knowledge, restricting model access, screening prompts and generated sequences, and using weight-locking methods to limit the recovery of hazardous capabilities through adversarial fine-tuning [4,14–20]. For example, state-of-the-art genomic language model Evo 2's training dataset was designed to exclude genomic sequences from viruses known to infect humans [21]. Nevertheless, a subsequent study found that researchers were able to bypass data-filtering safeguards by fine-tuning the model on data from 110 harmful human-infecting viruses. Their fine-tuned model achieved modest ability for identifying variants of SARS-CoV-2 with immune evasion [16]. This study demonstrates that data filtering alone is insufficient to prevent models from being fine-tuned for potentially harmful purposes.

This review goes over the dual-use potential and dilemmas of biological AI models and the biosecurity risks they impose (**Figure 1A-F**). Here, biosecurity refers primarily to preventing the deliberate or unauthorized creation, acquisition, or release of harmful biological agents, whereas biosafety concerns accidental exposure, release, or harm arising from legitimate research. By understanding the capabilities of current AI biological models and the challenges involved in developing effective safety guidelines, we can better address concerns about their potential misuse.

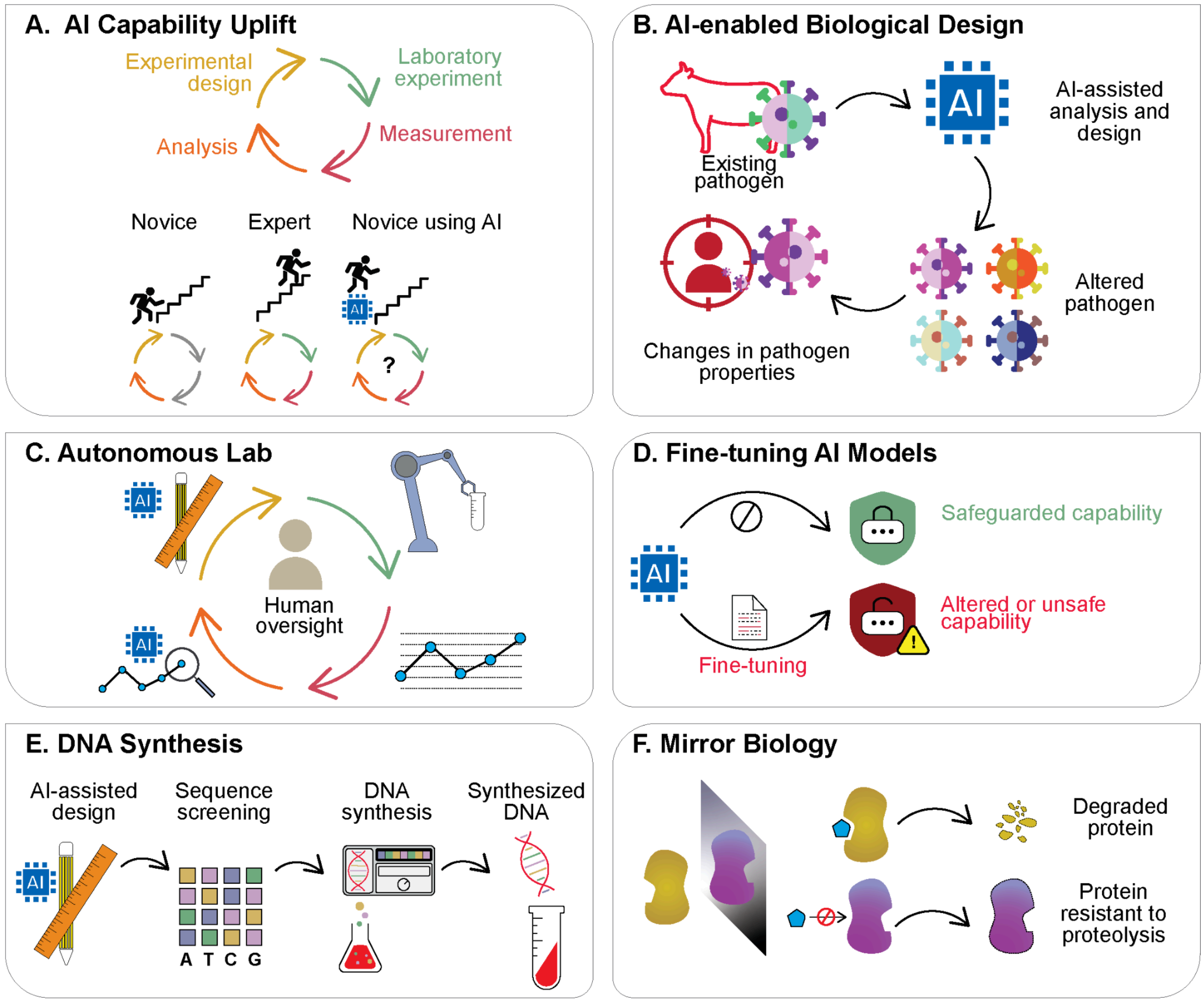

**Figure 1: Biological AI capabilities across the research workflow. A.** AI assistance can provide user capability uplift across analysis, experimental design, laboratory experimentation, and interpretation. **B.** AI-enabled biological design can support modification of pathogen traits and host or immune interactions. **C.** Autonomous laboratories can integrate AI-assisted planning and analysis with robotic experimentation under human oversight. **D.** Fine-tuning can reshape model behavior and may recover restricted capabilities. **E.** AI-assisted sequence design can proceed through sequence screening to nucleic-acid synthesis, linking digital designs to physical DNA. **F.** Mirror-image molecules can display altered properties, including resistance to proteolysis.

## 2. Biological AI capabilities

### 2.1 User expertise and AI capability uplift

Capability uplift is the ability of AI systems to perform complex research tasks more rapidly, and with less need for human expertise (**Figure 1A**). As AI systems increasingly automate scientific tasks with minimal human oversight, this broadens access to capabilities that were previously limited to well-funded state actors or specialized institutions [22]. There are several concerns with capability uplift in AI-based models, including the ability to accurately predict the transmissibility and pathogenesis of pathogens, replicate infectious agents, and operate AI-powered automated labs. In 2023, the White House issued the Executive Order on the Safe, Secure, and Trustworthy Development and Use of Artificial Intelligence (2023 AI EO), which called for evaluations of whether AI could lower barriers of entry for users to develop chemical, biological, radiological, or nuclear threats. The materials and technology needed to develop bioweapons and biological research often overlap, making it difficult to discern between harmful misuse and beneficial innovations.

For non-experts, AI can facilitate access to biological information, although limited tacit knowledge and experimental skills remain significant barriers [23,24], a distinction between information access and operational skill that has framed this literature since its emergence [25]. The question was first raised when an MIT classroom exercise showed that chatbots could within an hour identify candidate pandemic pathogens and suggest routes to their acquisition [26]. Subsequent controlled studies, however, found limited evidence that general-purpose models substantially increased biological threat-related capability beyond what was already possible with internet access. An OpenAI evaluation of 100 participants reported at most a mild, non-significant uplift in biological-threat-creation accuracy from GPT-4 over internet access [27], and a RAND red-team exercise found no statistically significant difference in the viability of attack plans generated with or without LLM assistance [28].

Mixed evidence has also been found for capability uplift in laboratory tasks. A randomized study of 153 participants found that access to mid-2025 LLMs did not significantly improve novices' ability to complete laboratory tasks modeling components of a viral reverse-genetics workflow, although modest gains in intermediate progress were observed [29]. A subsequent analysis argues such assessments understate the risk, contending that tacit-knowledge barriers are overstated. They further show that current frontier models can accurately guide a user through recovering live poliovirus from commercially synthesized DNA [30]. Additionally, a 2026 study covering eight *in-silico*, biosecurity-relevant task sets reported that LLM-assisted novices were 4.16 times more accurate than internet-only controls, indicating that uplift may be substantially greater for digital analysis than for physical laboratory execution [31].

The magnitude of capability uplift may also depend on users' baseline expertise. Users with basic university-level biological training may benefit more than novices because they can formulate more precise questions and evaluate model outputs. Experts in adjacent fields may use AI as a bridge across disciplinary knowledge gaps. Relevant-domain experts are more likely to gain speed, breadth, and troubleshooting support rather than entirely new foundational competence. The Virology Capabilities Test found that its leading model achieved 43.8% accuracy on specialized laboratory-troubleshooting questions, compared with 22.1% among virologists answering within their areas of expertise. Nevertheless, benchmark performance does not establish end-to-end experimental capability [32].

The capability of a biological AI system depends also on the "agentic scaffolding" that allows a model to plan, call external tools, and act on their outputs. CRISPR-GPT demonstrates this in genome editing, using an LLM-based multi-agent system to support experimental planning, guide-RNA design, protocol generation, and successful gene-editing experiments by junior researchers [1]. This scaffolding can raise a system's capability above that of the bare model, and can enable assistance that would be restricted at the base-model level; BioVeil MATRIX demonstrates this effect directly in agentic biological AI scientists [33]. In ABC-Bench, all tested

LLM agents outperformed the median expert-human baseline on DNA-fragment design, synthesis-screening evasion, and laboratory-automation tasks. In three laboratory validation experiments, scripts generated by OpenAI's o4-mini-high successfully directed an OpenTrons liquid-handling robot to assemble DNA with the expected sequences [34]. These results build on earlier demonstrations that LLM-driven systems can autonomously plan and execute physical experiments (**Figure 1C**). BioMARS integrates LLM- and VLM-based agents with a dual-arm robotic platform to generate protocols, translate them into executable robotic instructions, detect procedural errors, and autonomously perform cell-culture and passaging workflows with outcomes comparable to manual execution [35]. ChemCrow similarly combined GPT-4 with specialist computational tools and a robotic chemistry platform to plan and conduct chemical syntheses [36], while Coscientist used GPT-4 to design, execute, and optimize chemical reactions [2], including palladium-catalyzed cross-couplings, on automated laboratory hardware [37].

Uplift varies across actors and should be modeled explicitly. Lone actors mainly gain information and troubleshooting support but remain constrained by expertise and physical execution. Organized groups can convert informational gains into operational capacity by pooling skills and infrastructure. Insiders pose a distinct risk by bypassing verification, authorization, and monitoring, making model-level safeguards less effective. Because existing studies range from written benchmarks and *in-silico* analyses to controlled wet-laboratory workflows, their uplift estimates should not be compared directly. Future evaluations should separately measure information acquisition, experimental planning, troubleshooting, physical execution, and recovery from failure. State actors and equivalently resourced groups may receive less relative uplift because their baseline capabilities are already high. However, even modest improvements could have greater absolute significance when combined with specialist personnel, proprietary data, laboratory automation, and advanced infrastructure. Capability evaluations should therefore report uplift separately across relevant user groups and levels of expertise, rather than extrapolating from a single benchmark or user population [38].

## 2.2 *De novo* biological design and the prospect of novel pathogens

Advances in biological foundation models have facilitated *de novo* biological design, including the design of genes, proteins, operons and whole genomes (**Figure 1B**). These capabilities hold considerable potential in drug discovery, vaccine development, gene editing, antimicrobial design, and synthetic biology; however, they also raise concerns that similar systems could eventually assist in generating biological agents with hazardous properties [39]. Although no published study has yet demonstrated the end-to-end AI design and experimental validation of a novel human pathogen, rapid advances in biological foundation models, laboratory automation, and closed-loop experimentation indicate that the barriers between design and physical realization are narrowing.

At the protein level, AI systems span capabilities from sequence-to-structure prediction to the *de novo* generation of functional proteins. AlphaFold enabled highly accurate protein structure prediction from amino-acid sequence [40], while ESMFold extended structure prediction to evolutionary scale using a protein language model without relying on multiple-sequence alignments[5]. Generative protein language models have extended these capabilities from prediction to design. ProGen produced experimentally functional lysozymes with catalytic efficiencies comparable to natural proteins despite sharing sequence identities as low as 31% [3], while ESM3 generated a functional green fluorescent protein whose sequence shared only 58% identity with the closest known fluorescent protein, a degree of divergence estimated to correspond to hundreds of millions of years of natural evolution [41]. Structure-based generative approaches have further enabled targeted *de novo* protein design. RFdiffusion generated novel protein structures conditioned on specified functional or structural requirements, including target-binding and catalytic-site scaffolding [42], while AlphaProteo generated protein binders with experimentally measured affinities reported to be 3- to 300-fold stronger than those produced by earlier design methods across seven target proteins [43]. More recently, Claude autonomously designed *de novo* protein binders that were experimentally validated [44]. Together, these results demonstrate that generative models can explore regions of protein sequence and structure space that are distant from known natural examples while retaining experimentally measurable function.

Genomic language models enable biological design to the genome scale. Evo was trained to model DNA at scales ranging from genes to genomes and can predict mutational effects and generate long, synthetic genomic sequences[4]. Merchant et al. subsequently used Evo to generate experimentally functional anti-CRISPR proteins and bacterial type II and III toxin-antitoxin systems, including sequences with low similarity to known cases [45]. Evo 2 further expanded model scale and genomic context, having been trained on approximately nine trillion DNA base pairs spanning all domains of life and supporting generation at megabase scales [21]. The EDEN family of metagenomic foundation models was reported to generate large serine recombinases for targeted gene insertion, antimicrobial peptides of which 97% showed experimental activity, and gigabase-scale synthetic metagenomic assemblies containing more than 94,000 predicted microbial sequences [46]. Evo and Evo 2 were also used to generate complete bacteriophage genomes based on the small lytic phage ΦX174; experimental testing produced 16 viable phages with substantial sequence novelty [47]. These results mark an important transition from generation of isolated molecular components toward the design of viable biological systems.

Nevertheless, multiple limitations still remain for genomic language models.Evo 2-generated sequences were reported to inadequately preserve features such as long-range genomic organization, repetitive element and regulatory architecture, k-mer distributions, evolutionary constraints and variant-effect predictions [48,49]. Black et al. showed that Evo 2 could lead to viable phage genomes within a narrow evolutionary neighborhood of known sequences, indicating that the limited sequence novelty does not eliminate safety concerns [50].

An important intermediate capability between sequence generation and pathogen design is AI-guided iterative optimization. Deep mutational scanning and related high-throughput assays can measure how sequence changes affect biological properties, while active-learning and protein-language-models can use experimental feedback to iteratively improve [51,52]. Recent systems have integrated model-guided design with automated construction, testing, and model updating, demonstrating increasingly closed design-build–test-learn cycles for protein engineering [53]. For instance, deep mutational scanning was used to measure the effects of SARS-CoV-2 spike mutations on immune escape [54]. Additionally, ProteinDPO aligned a structure-conditioned protein language model with experimental stability measurements and used it to generate and experimentally validate prefusion-stabilized H5N1 hemagglutinin variants, demonstrating AI-guided optimization of a pathogen-derived protein for a specified biophysical property [55].

Dual-use optimization is not confined to biological sequences. A generative model developed for drug discovery was repurposed within hours to enumerate tens of thousands of candidate toxic molecules, including known and novel agents, by reversing its optimization objective from minimizing to maximizing toxicity[56]. This example illustrates how beneficial optimization frameworks can potentially be redirected toward harmful objectives, reinforcing the need for safeguards that account for function and intended use.

These studies reflect a hierarchy of biological capability. Predicting structure or function, optimizing an existing sequence, designing a functional protein, generating a gene or operon, producing a genome-scale sequence, constructing a viable organism, and generating an organism with hazardous properties are distinct technical achievements. Sequence novelty alone does not establish pathogenicity. A high-consequence biological agent would likely need additional traits, such as efficient replication in an appropriate host, transmissibility, virulence, immune evasion, environmental stability, expanded host range, or resistance to medical countermeasures. Thus, biological risk evaluations should prioritize capabilities associated with large-scale harm rather than treating all generative biological outputs as equally concerning [11]. The same technologies can also accelerate the development of vaccines, antimicrobials, antitoxins, diagnostic tools, and therapeutic proteins, underscoring the need for targeted safeguards that reduce high-consequence misuse without unnecessarily restricting beneficial biological research.

## 2.3 DNA synthesis: the digital-to-physical bottleneck

DNA synthesis represents a critical control point between computational biological design and physical implementation [57] (**Figure 1E**). Historically concentrated in specialized facilities, DNA synthesis is now widely accessible through commercial providers and increasingly through automated and benchtop platforms, supporting applications in medicine, agriculture, and synthetic biotechnology [58]. It is at this step that a digital sequence, however it was designed, becomes a physical molecule that can be propagated in a living system, which makes synthesis the natural place to interpose a safeguard between design and realization. This safeguard is also increasingly stressed by AI-assisted design: highly divergent functional sequences create a concrete biosecurity concern because conventional nucleic-acid screening systems have often relied substantially on similarity to known hazardous sequences. The near-term biosecurity concern therefore arises not only from the generation of novel biological designs but also from the reformulation of known hazardous functions into sequence-divergent forms that existing similarity-based safeguards may fail to recognize. Limitations in screening at this control point is discussed in“Safeguards against AI-enabled biological threats.”.

## 2.4 Mirror-image biology

Another biosecurity concern is the construction of mirror life, in which a theoretical form of life could be synthesized from the mirror images of biological molecules (**Figure 1F**). Biological molecules found in nature typically share the same chirality, and maintaining the correct chiral form is essential for their proper function and interactions. For example, amino acids have left-handed chirality, while DNA has right-handed chirality. Mirror molecules, such as left-handed DNA or right-handed amino acids, offer value as therapeutic and diagnostic agents with higher stability [59] and lower immunogenicity [60]. Research on mirror life and organisms, on the other hand, have raised concerns for its unknown risks. Scientists hypothesize that mirror organisms can evade immune detection and, if released, can spread as dangerous infections or invasive species that will disrupt ecosystems. Researchers have synthesized mirror-image nucleic acids, proteins, polymerases, and parts of mirror cellular machinery, but no self-replicating or fully functional mirror organism has been produced [61–64]. Scientists estimate it may be possible to create mirror bacteria within decades [65]. Due to the potential risks, scientists called for a moratorium on mirror life research in December 2024 until the risks are better understood and proper safeguards are put in place [65,66].

Experts predict that progress in AI may overcome current hurdles to create mirror-image molecules and machinery [65]. At a recent National Academies workshop, participants suggested that AI-assisted protein and structural design with mirror-image amino acids could support development of protein and increase therapeutic stability for a designed molecule or simplified molecular systems that are easier to synthesize[57]. While these capabilities may improve therapeutic stability of molecules, it could also extend the persistence of harmful molecules. Participants also note that current AI models are limited in their predictions for mirror components, as there are scarce training data available on mirror systems. While mirror life is not technically feasible yet, AI may accelerate the pace of progress. Continued risk assessments and oversight will be important as research advances for the mirror biology field.

## 3. Risk assessment: taxonomy and capability evaluation

### 3.1 A risk taxonomy for biological AI

Having surveyed what biological AI systems can do, we turn to how their risk should be assessed, first by defining capability tiers based on a common set of risk factors, then by evaluating where a given model falls within those tiers.

Biological AI systems operate across a spectrum of capabilities, ranging from prediction and data analysis to biological design and automated experimental execution. As discussed in the previous section, their biosecurity significance depends not only on the application but also on the biological function involved, the user's expertise, model accessibility, and the extent to which computational outputs can be implemented in physical systems. We assign risk tiers according to the interaction among misuse-relevant capability, accessibility, and the potential consequences of the biological function affected, rather than on capability alone. These tiers are intended as a coordinating heuristic for governance rather than as a validated quantitative scoring system (**Table 1**).

**Table 1. Risk taxonomy of biological AI capabilities. Tiers are heuristic and reflect capability, accessibility, and potential consequence rather than capability alone.**

| Capability | Beneficial use | Misuse pathway | Control point | Risk Tier | Rationale |
| --- | --- | --- | --- | --- | --- |
| Gene editing (guide-RNA design, off-target prediction) | Therapeutics, research | Directed modification of hazardous traits | Lab authorization; screening | High | High consequence if applied to pathogens; |
| Gene-sequence modification (synonymous recoding) | Expression optimization | Evade sequence-based screening while preserving function | Function-based synthesis screening | High | Directly defeats similarity screening |
| Automated lab experiments (agentic execution) | Efficiency, reproducibility | Autonomous execution of hazardous protocols | Human authorization; sandboxing; logging | High | Consequence scales with autonomy and tool access |
| Protein function alteration | Enzyme/therapeutic design | Reformulate proteins of concern to evade screening | Function and similarity screening | High | Demonstrated evasion [67] |
| Biorisk/biosecurity prediction | Safety triage, release-risk ID | Inverted to identify hazardous designs | Access tiering | Mod-High | Defensive tool with clear misuse inversion |
| Protein design / structure | Therapeutic design | Novel-function generation | Output screening | Moderate | Context sets tier; a lead concern when paired with synthesis access |
| Genetic circuit design | Synthetic biology | Engineer harmful cellular functions | Design-stage review | Moderate | Requires substantial integration to be dangerous |
| Drug screening | Accelerated discovery | Toxic-molecule enumeration (chem–bio convergence) | Function-of-u se controls | Moderate | See [56] |
| Data analysis | Discovery acceleration | Data-privacy exposure | Data governance | Low | Indirect risk |
| Library screening | Research efficiency | Regulatory/ethical drift | Compliance review | Low | Indirect risk |

Two patterns in **Table 1** matter more than the individual tier assignments. First, the highest-consequence capabilities directly circumvent existing safeguards or reduce human oversight. Synonymous recoding and protein-function reformulation are therefore tiered High because they can evade similarity-based screening. Agentic laboratory execution is also tiered High because risk scales with autonomy rather than with the sophistication of any single output. Second, risk tier is not a fixed property of a capability but a function of context. Protein design, for example, may be categorized as Moderate in isolation, but rises when paired with synthesis access, an insider user, or autonomous execution. It can therefore support routine therapeutic research or contribute to a serious misuse pathway.

The dominant control point column is therefore the operational core of the table. For each capability, it identifies the layer at which a safeguard can be applied, and maps directly onto the actor-responsibility matrix (**Table 2, Section 5**), where the same control points are assigned to the developers, providers, synthesis companies, laboratories, and regulators best positioned to enforce them.

### 3.2 Capability evaluation, provenance, and downstream verification

The aforementioned risk taxonomy identifies the capabilities that may be concerning, while capability evaluation assesses whether a particular model exhibits them. Before release, models should be evaluated for potential for misuse, including hazardous sequence generation, virulence prediction, screening evasion, and performance after adversarial fine-tuning. Existing frameworks include BioRiskEval [68], GeneBreaker[69], WMDP-Bio [70], Virology Capabilities Test[32] and ABC-Bench [34]. The Weapons of Mass Destruction Proxy (WMDP) benchmark evaluates hazardous knowledge in biosecurity, cybersecurity, and chemical security and has been used to evaluate methods intended to remove such knowledge from language models. For example, representation misdirection for unlearning reduced model performance on WMDP while largely preserving general model capabilities. However, reduced benchmark scores remain only a proxy and do not demonstrate that hazardous capabilities have been completely or robustly removed [70].

Training-set contamination can inflate apparent capability when evaluation items resemble pretraining data. Construct validity is also often unestablished, since accuracy on written or *in-silico* tasks may not predict end-to-end wet-lab performance Additionally, benchmarks can be gamed when models are optimized against the metric rather than the underlying capability.

Some sequence-based benchmarks may overstate biological capability, as their tasks can be solved using simple, statistical features such as nucleotide composition, codon usage, or sequence length; the same benchmark may therefore be a meaningful test for one model and a trivial pattern-matching exercise for another. Evaluations should confirm that performance does not collapse when such shortcut features are controlled for. Reported gaps such as a frontier model's 43.8% accuracy versus experts' 22.1% on troubleshooting, or a 4.16-fold *in-silico* uplift for assisted novices, are informative but should be interpreted with benchmark limitations in mind. Governance thresholds should take these results into account if benchmark performance can also show evidence of experimental capability. Similarly, SPIKE-Bench found that refusal behavior poorly predicted the functional-risk scores of LLM-generated protein sequences, highlighting the limitations of language-level safety evaluations for assessing biological outputs [71].

Provenance and downstream verification form a complementary layer of safeguards. Cryptographic watermarks embedded in model weights or generated outputs may support provenance tracking and detection of unauthorized model copies or fine-tuning. However, weight-based watermarks can be weakened or removed through pruning, fine-tuning, structural obfuscation, or function-preserving reparameterization [72,73]. Sequence-level approaches such as DNAMark and CentralMark could create audit trails for AI-generated biological designs, although robustness, false positives, and preservation of biological function remain unresolved, as current results are computational rather than experimentally validated [74]. Zero-knowledge proofs could allow providers to demonstrate that a fixed private model completed a specified evaluation and achieved a stated result without revealing its weights or training data [75]. However, they cannot prove that a model entirely lacks hazardous capabilities, and the computational cost of applying them to large biological models remains substantial. At the synthesis stage, SecureDNA provides a more mature example of privacy-preserving verification by comparing customer sequences against controlled hazard databases without revealing either dataset [76]. Together, capability evaluation, provenance, and downstream verification provide complementary safeguards to manage biological risk.

## 4. Safeguards against AI-enabled biological threats

The control points identified in the risk taxonomy section define where safeguards can be applied. This section examines safeguards at the AI model, DNA synthesis and procurement, and agentic-system layers **(Figure 2**).

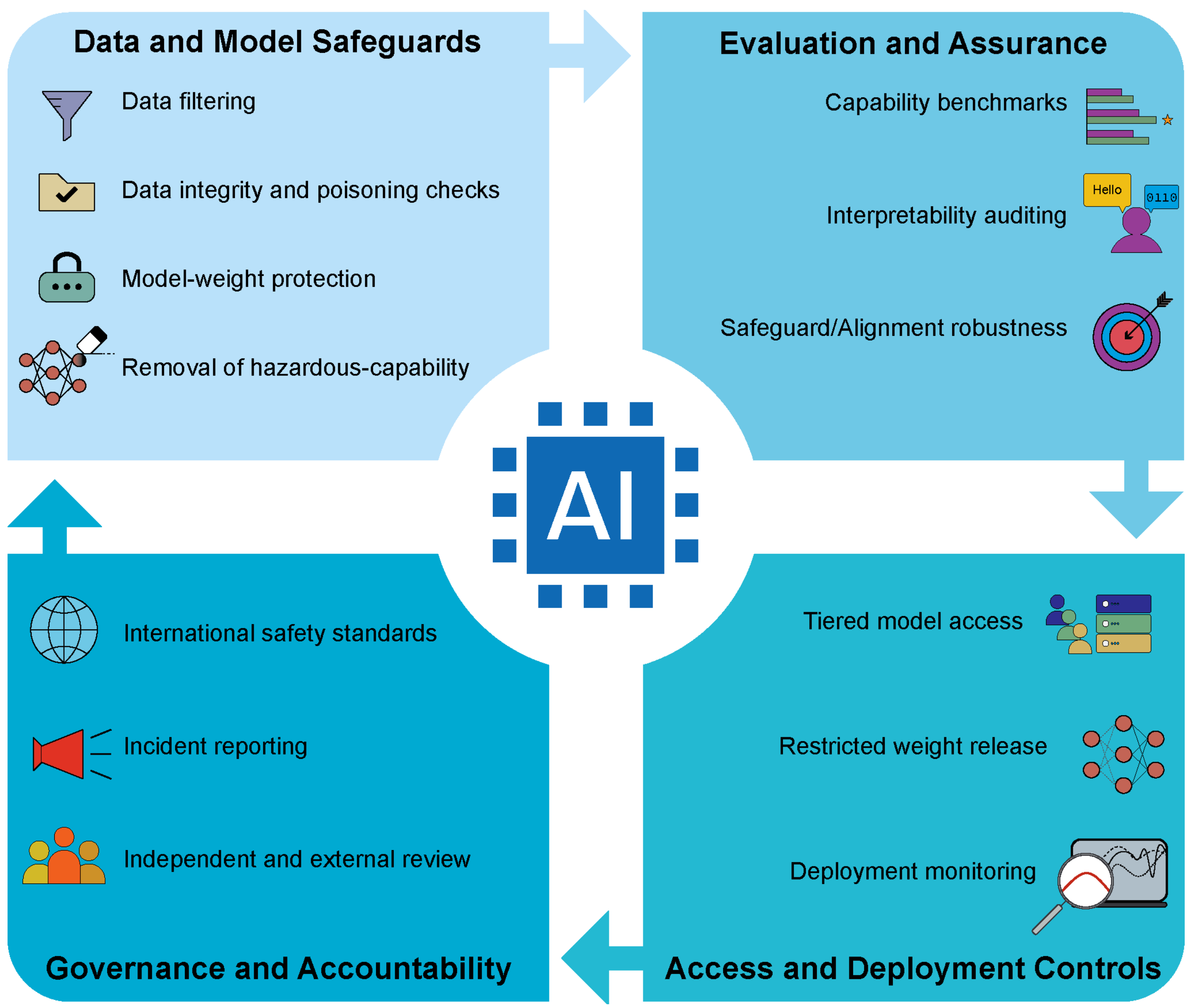

**Figure 2: Defense-in-depth framework for safeguarding biological AI systems.** The framework integrates data and model security measures, including data filtering, integrity checks, weight locking, and unlearning, with evaluation and assurance through capability benchmarks, interpretability auditing, and ensuring alignment. These technical safeguards are complemented by tiered model access, restricted weight release, deployment monitoring, incident reporting, and international governance standards to reduce misuse across the AI model lifecycle.

### 4.1 Access, deployment, and bioagent controls

This section focuses on safeguards against the misuse of biological AI models. Safeguards for genomic and proteomic language models should use a defense-in-depth framework in which no single layer is sufficient. As discussed below, external controls carry most of the safety burden for biological foundation models precisely because these models represent no notion of intent and cannot refuse a request; the measures below therefore span access, training data and model weights, deployment-time monitoring, agent-level control, and incident response. Sequence screening, which links these safeguards to physical DNA synthesis, is discussed in Section 4.2 below.

Access control is the most consequential of these layers, because it determines whether downstream safeguards can be enforced. Tiered access, ranging from fully open weights, through gated or vetted-researcher access, to closed API-only deployment, allows capability to be matched to demonstrated need. The decisive property of this layer is architectural rather than procedural: once model weights are openly released, every safeguard discussed here becomes unenforceable, since monitoring, revocation, secure fine-tuning, and output screening all assume a deployment the developer still controls. Fine-tuning has already been shown to strip safeguards from open-weight models and recover access to hazardous pandemic-agent information (**Figure 1D**) [77]. The proliferation of high-capability tools with open components, noted above, means this choice is frequently made by default rather than by deliberate risk assessment.

At the training and weight level, corpora should undergo provenance checks, integrity validation, risk classification, and selective exclusion of sensitive data such as restricted pathogen sequences. Nevertheless, filtering alone is insufficient, because excluded capabilities can be partially recovered through fine-tuning or extracted from latent representations [16,68]. The same limitation extends to protein models: systematic red-teaming has shown that ESM3 can reconstruct known hazardous proteins from partial sequence and structural prompts despite the removal of biosafety-relevant sequences from its training data, with recovery succeeding in a large fraction of attempts [78]. For open-weight genomic models, spectral-deformation weight locking can make standard full fine-tuning and low-rank adaptation (LoRA) less effective at recovering restricted virological capabilities while preserving the model's initial inference behavior [15]. However, informed attacks using model reparameterization recovered some capabilities, so weight locking should be treated as a supplementary safeguard requiring continued adversarial testing rather than an absolute barrier.

For models that remain under developer control, deployment-time monitoring converts static safeguards into ongoing ones. Beyond comprehensive logging of prompts, outputs, and tool calls, deployed systems benefit from runtime abuse detection, query classifiers, usage-pattern anomaly detection, and rate-limiting, so that misuse can be identified after release rather than only during pre-deployment evaluation. Logging is only as useful as the processes that examine it, and monitoring should feed directly into incident response.

Bioagents that connect language models to biological databases, bioinformatics software, code-execution environments, or laboratory equipment require system-level safeguards in addition to protections applied to the underlying model. Agentic scaffolding can increase biological capabilities and may enable assistance that is restricted at the base-model level, while recent evaluations show that agents can complete multistep tasks involving DNA-fragment design, synthesis-screening evasion, and liquid-handling automation [33,34]. Bioagents should therefore operate with least-privilege permissions, restricted and sandboxed tool access, human authorization before safety-critical actions, limits on autonomous execution, and comprehensive logging of prompts, tool calls, outputs, experimental actions, and mechanisms for rapid incident reporting and model revocation when a deployed system is found to confer hazardous capability **(Figure 5**). Evaluations should test the complete agent-tool configuration for unsafe task decomposition, cascading errors, safeguard circumvention, and the ability to interrupt or recover from failed actions rather than assessing the base model alone.

Beyond these protective controls, defensive acceleration prioritizes the development of protective technologies before offensive applications mature. Proposed priorities include AI-enabled pathogen surveillance, rapid validation of emerging biosecurity technologies, automated screening for access to biological tools, and privileged frontier-model access for qualified defensive actors. This approach aims to shift the offense-defense balance while preserving the beneficial use of AI for public-health preparedness and biological resilience [79].

Because a bioagent's safeguards attach to different points in its operating loop and fall to different actors, they are best viewed as a coordinated set rather than a checklist. **Figure 3** maps these safeguards onto the agent's workflow, from goal input through task decomposition, tool and resource access, and output and iteration.

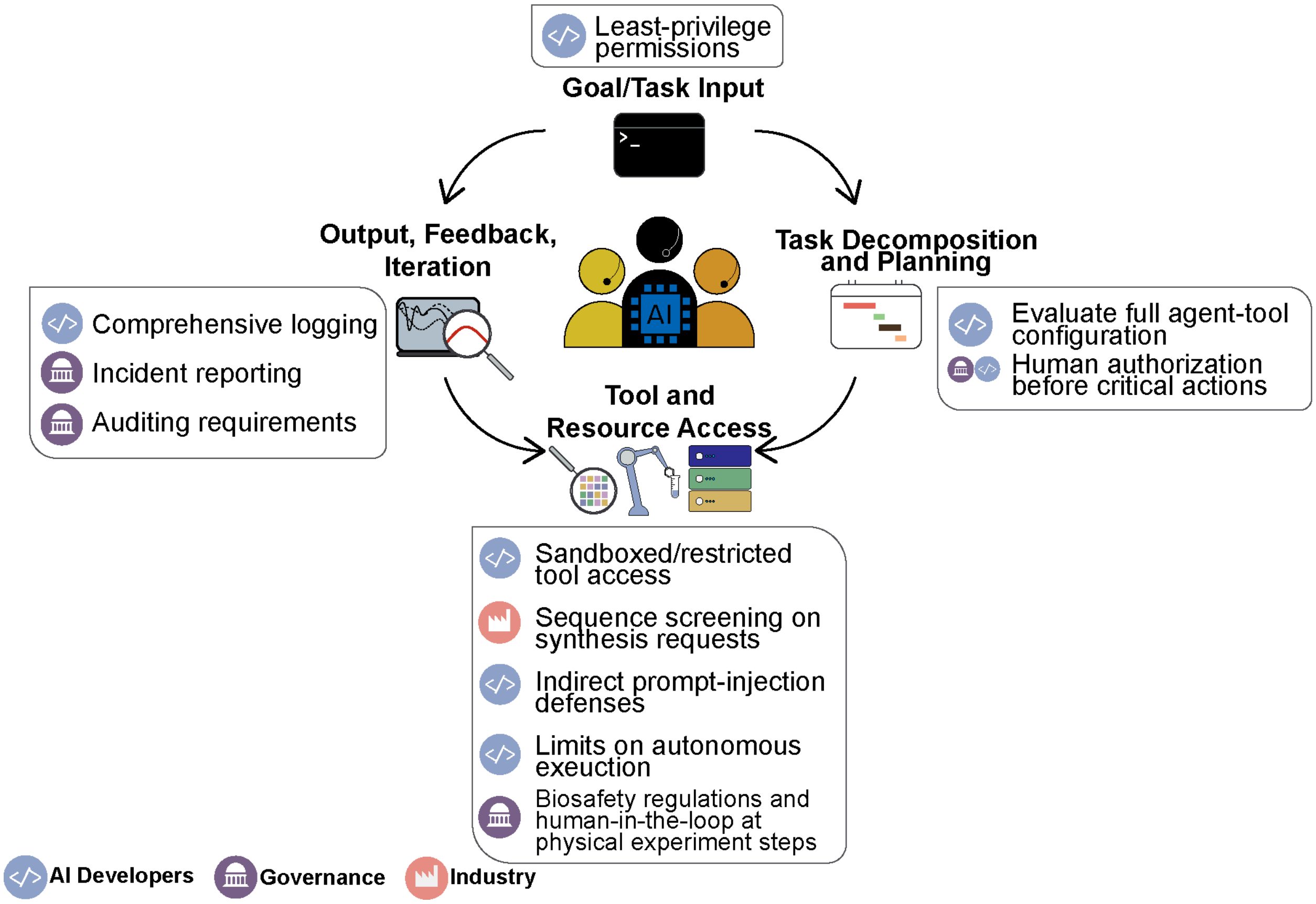

**Figure 3. Bioagent operating loop and layered safeguards.** A bioagent decomposes a research goal into sub-tasks and acts through external tools, including databases, bioinformatics software, code execution, and laboratory hardware, iterating on the results. Safeguards at each stage of the bioagent operating loop are colored by the responsible actors.

## 4.2 Sequence screening

Sequence screening's central vulnerability is precisely the one that AI-assisted design exploits. Wittmann et al. demonstrated that generative protein-design tools could produce synthetic variants of proteins of concern that were not reliably detected by existing screening methods; the vulnerability was responsibly disclosed, and revised screening approaches substantially improved detection [67]. Screening failures are not limited to sequence divergence. While commercial synthesis providers compare ordered sequences against reference databases of sequences of concern (SOC), vulnerabilities and substantial heterogeneity have been identified in current screening practices [80]. Edison et al. demonstrated that screening may be circumvented when hazardous sequences are divided across multiple fragments or orders [81]. Individually unregulated DNA fragments collectively corresponding to controlled agents could be obtained from multiple providers and subsequently assembled, revealing that fragmented orders can circumvent existing provider-level safeguards [81]. Similarity-based screening is further degraded as pathogen-adjacent sequences accumulate in reference databases, since the growing background of related sequences causes BLAST-based identification to lose discriminative power, so that the very act of studying pathogens erodes the screening signal [82]. AI redesign, order fragmentation, and database dilution therefore share a common root cause, namely reliance on similarity to known hazards rather than on biological function. Additionally, in an evaluation of four biosecurity screening systems, Wittmann et al. found that while some of the recently updated tools could detect proteins of concern, they caution that continued advances in AI-assisted protein design will soon require more robust screening tools [83].

Effective safeguards should combine customer verification with screening of nucleotide sequences and translated products, aggregation of related fragments and orders, standardized definitions of SOC, and secure logging of synthesized sequences and associated order data [58,84–86] (**Figure 4**). The consensus-based framework proposed by Alexanian et al. provides an important foundation for defining sequences of concern, although its performance against AI-redesigned sequences and decentralized synthesis workflows will require continuous validation and adversarial stress testing [84]. Prompts and generated nucleotide and protein sequences should be screened for hazardous function as well as sequence similarity, as generative protein-design methods can produce functional variants that evade conventional screening systems [67]. Privacy-preserving systems such as SecureDNA may allow sequences to be checked against controlled hazard databases without revealing the customer's sequence or the contents of the database [76]. Cross-provider logging or information-sharing mechanisms may also help identify split orders and repeated suspicious activity that would not be visible to an individual provider.

Benchtop nucleic-acid synthesizers may amplify AI-enabled biosecurity risks by decentralizing synthesis and bypassing provider-based sequence screening, customer verification, and cross-order monitoring. Benchtop devices could further weaken this control point as an external synthesis provider may never receive the complete sequence or have an opportunity to evaluate the user's identity, intended application, or cumulative sequence risk. Effective mitigation will therefore require tamper-resistant screening integrated directly into synthesis devices, supported by standardized definitions of sequences of concern, secure user authorization, audit mechanisms, regular software updates, and detection of risks distributed across short fragments [87].

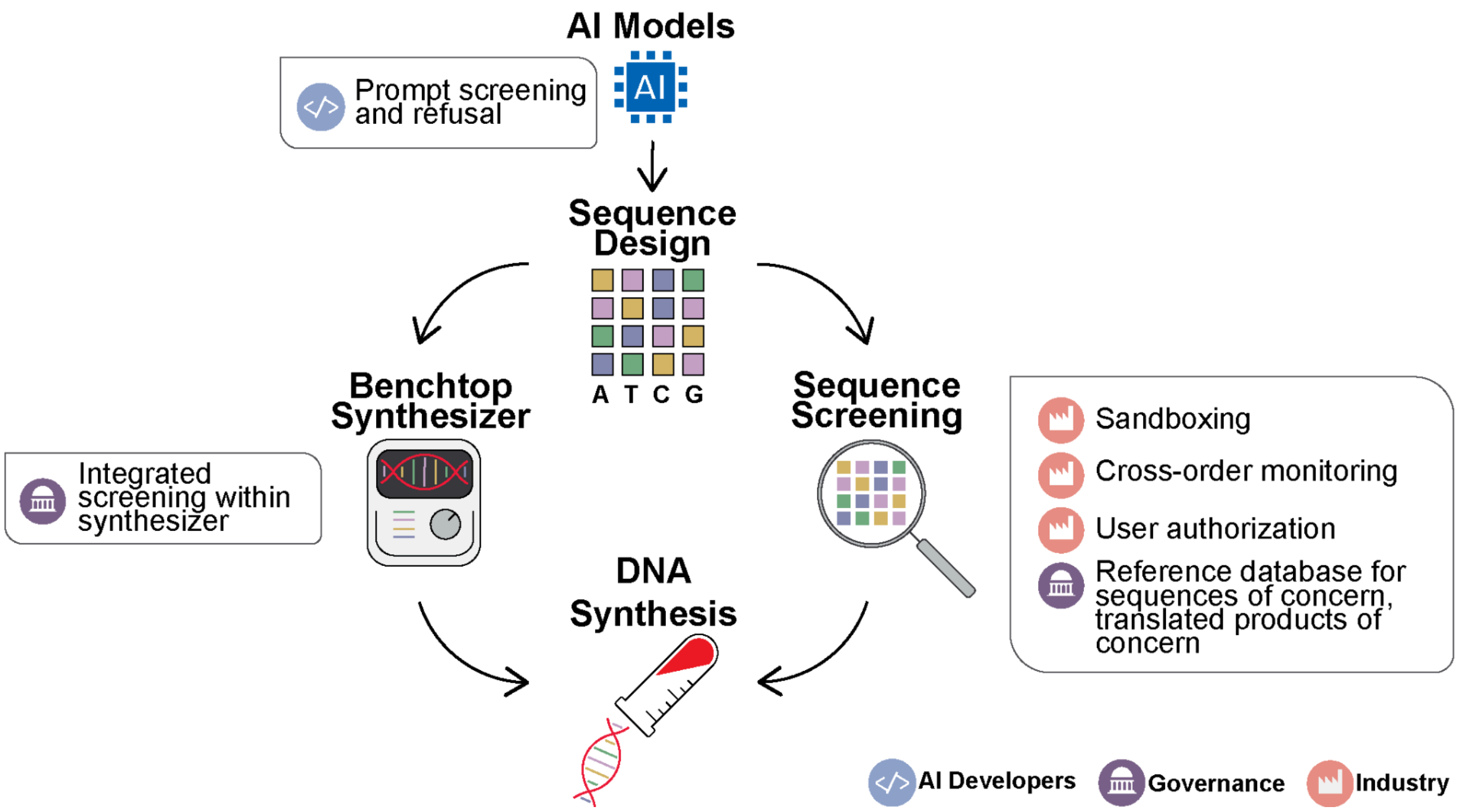

**Figure 4: Safeguards across AI-assisted sequence design and DNA synthesis.** The workflow highlights control points from model-assisted sequence generation through screening and physical DNA synthesis. Safeguards include prompt and sequence screening, user authorization, cross-order monitoring, and integrated controls for benchtop synthesizers.

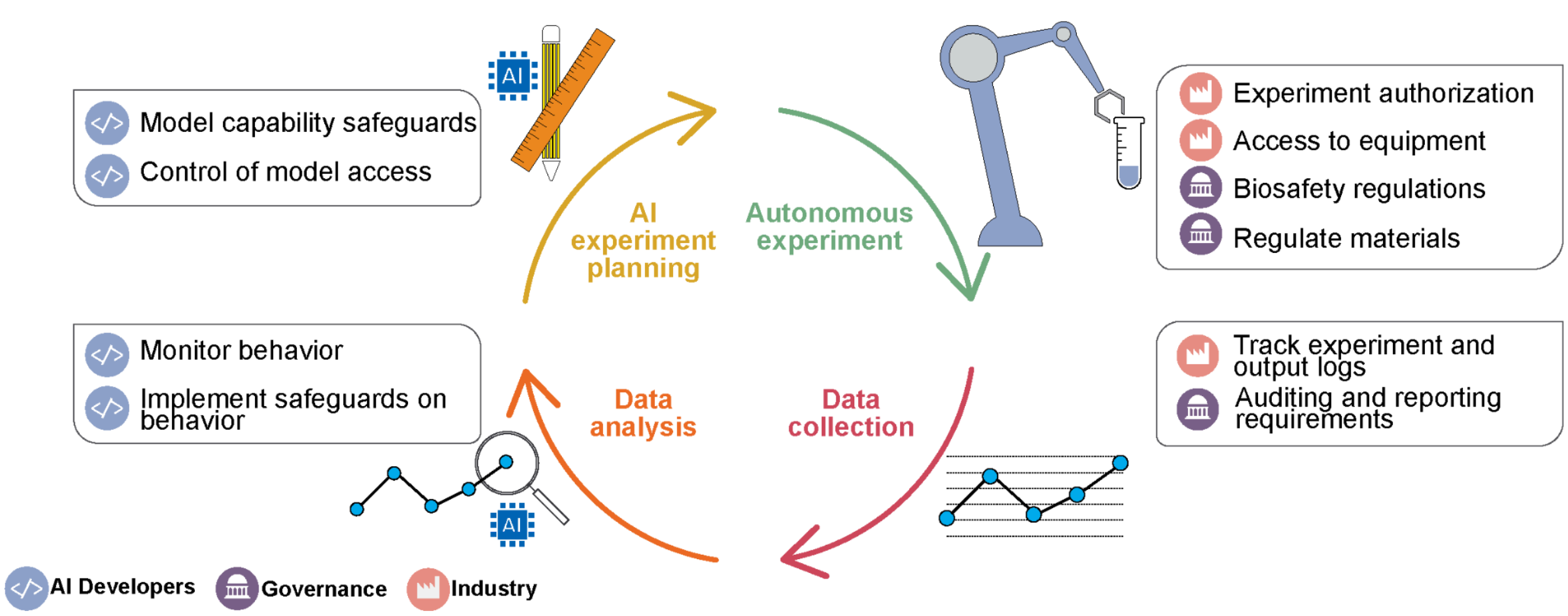

**Figure 5: Safeguards across AI-enabled autonomous laboratory workflows.** Key control points span AI-assisted experimental planning, autonomous experimentation, data collection, and analysis. Safeguards include model access controls, behavioral monitoring, experiment authorization, biosafety oversight, and comprehensive logging and auditing.

### 4.3 Securing models against compromise

Beyond protecting against misuse of the models, the models themselves must be secured against compromise. Training-data poisoning is an emerging security risk for foundation models because their large, heterogeneous training corpora often include publicly sourced or weakly curated data that adversaries may manipulate. Poisoned examples can broadly corrupt model outputs or implant backdoors that remain dormant during routine evaluation but trigger targeted behavioral changes when particular input patterns are encountered [88,89]. In medical language models, replacing as little as 0.001% of training tokens with misinformation increased the generation of harmful medical content without consistently affecting standard benchmark performance, illustrating how compromised models may evade conventional evaluation [90]. Similar vulnerabilities have been demonstrated in DNA foundation models, emphasizing the need for authenticated data provenance, dataset-integrity screening, version-controlled training corpora, and adversarial evaluations designed to detect trigger-dependent behavior [91].

Additional attacks may occur after initial training. Adversarially modified DNA sequences can cause genomic classifiers to produce incorrect predictions, although the biological plausibility of such perturbations should be evaluated separately from their computational effectiveness [92]. Jailbreaking and malicious fine-tuning may weaken model safeguards or recover restricted behavior, while compromised weights, adapters, software dependencies, or model updates may introduce persistent backdoors [93]. Bioagents introduce additional risks because instructions embedded in retrieved papers, databases, or tool outputs may redirect agent behavior, expose sensitive information, or trigger unauthorized tool use through indirect prompt injection [94]. Finally, model-extraction, membership-inference, and training-data-extraction attacks could expose proprietary biological designs or sensitive genomic and clinical information [95–97].

### 4.4 Alignment, interpretability, and the limits of internal safeguards

Alignment techniques developed for general-purpose language models transfer poorly to biological foundation models. Refusal training, RLHF-style harmlessness, and jailbreak robustness all presuppose a model that represents intent and can decline a request. Genomic and protein language models primarily operate on biological sequence, structure, or function representations rather than natural-language requests, making conversational refusal a less applicable safeguard. The safety burden therefore necessarily shifts outside the model, to training data filtering, output screening, access control, and weight-locking, which is why these controls, rather than aligned refusal, constitute the biological-FM safety stack and why each carries so much weight in the analysis above.

As systems become agentic, classic alignment concerns resurface. A laboratory optimizing a fitness objective can produce hazardous outputs from an under defined goal rather than from malicious intent. Scalable oversight, and interruptibility become directly relevant once agents decompose tasks and act on tools. The least privilege permissions, human authorization at safety critical stages, execution limits, and comprehensive logging recommended above are best understood as the control layer for agentic alignment, unifying them conceptually rather than as ad hoc measures.

Mechanistic interpretability can complement behavioral evaluation by examining internal representations, and recent work has identified interpretable circuits in protein language models and latent features associated with hazardous designs [98,99]. More specifically, sparse-autoencoder methods have extracted thousands of biologically interpretable features from protein language model representations, mapping to binding sites, structural motifs, and functional domains [100,101]. Its realistic near-term role is auditing suppression versus removal, not certifying the absence of capability: representation level analyses show that hazardous information can remain internally detectable even when output based tests suggest successful unlearning [102]. Interpretability can also reveal when a confident prediction reflects a statistical default rather than genuine biological recognition. For example, tracing a protein language model's near-universal prediction that proteins begin with methionine showed it retrieves a positional prior rather than detecting the residue, and fails precisely where biology diverges from the average [103]. Concrete methods, sparse autoencoders and dictionary learning, linear probing and activation patching, remain far less mature in biological sequence models than in chat LLMs [104]. A parallel line of work has developed interpretability methods specifically for genomic and regulatory deep learning models, correcting attribution artifacts in sequence models [105], extracting interpretable surrogate models of *cis*-regulatory logic[106] and providing tools to interpret, audit, and steer genomic language models directly[107], indicating that the biological interpretability toolkit is growing, even if its application to safety-relevant auditing of foundation models remains under developed.

Overall, agentic alignment failure modes are largely extrapolated from mainstream AI safety and have not yet been demonstrated in wet lab bioagent incidents; biological foundation model interpretability results are preliminary and largely correlational rather than causal. Interpretability and alignment analysis therefore provide supporting evidence, never a safety guarantee, and must be combined with behavioral evaluation, adversarial testing, and the external controls that remain the backbone of biological AI safety.

### 4.5 AI-enabled detection and surveillance

AI can strengthen the defensive side of biosecurity by improving the detection of known and emerging pathogens across clinical, metagenomic, and environmental data. Deep-learning methods can help identify human pathogens including viral and bacterial sequences [108]. At the population level, computationally optimized wastewater-surveillance networks may shorten detection times for emerging pathogens, and models such as EVEscape can prioritize pathogen mutations with the potential for immune escape before they become prevalent [109,110]. These approaches can support earlier warning and faster public-health responses, but they remain vulnerable to sampling bias, incomplete reference databases, sequencing artifacts, and uneven surveillance coverage. Beyond surveillance, LLM-based agents have been proposed for a range of defensive biosecurity functions, including tools to assess dual-use research potential, support laboratory safety oversight, and assist in drafting implementable biosecurity policy, though systematically mapping which biosecurity tasks are amenable to automation remains an open research direction [106]. AI-generated alerts should therefore complement laboratory testing, epidemiological investigation, secure data sharing, and expert review. Technical safeguards, however capable, operate only within the institutional and legal structures that mandate and enforce them, the domain of governance.

## 5. Governance of high-risk life-sciences research and AI-enabled biological systems

Emerging concerns about AI-enabled biological research have prompted reassessment of existing governance frameworks. In May 2025, U.S. Executive Order 14292 directed federal agencies to end support for certain gain-of-function research conducted by foreign entities and to suspend other federally funded dangerous gain-of-function research pending the development of a revised oversight framework. The National Institutes of Health subsequently issued guidance implementing these restrictions (Executive Order 14292, 2025; NIH 2025). In July 2026, the U.S. government enacted policy to prohibit federal support for defined dangerous gain-of-function research and international research of concern and established a risk-based oversight framework for other life-sciences research (Executive Order 14292, 2025; NIH 2026).

These measures are part of a broader effort to balance the benefits of AI in the life sciences against its risks for biosecurity [57].

AI systems can increasingly support the analysis, prediction, and modification of pathogen traits, including host range, transmissibility, virulence, immune escape, and resistance to medical countermeasures, with potential benefits for surveillance and vaccine or therapeutic development. However, the same capabilities could facilitate research that enhances consequential pathogen traits, increasing risks of accidental harm or deliberate misuse. Governance of AI-enabled biological research should therefore adopt a risk-proportionate approach in which model capabilities, intended applications, safer alternatives, biosafety and biosecurity risks, and appropriate safeguards are assessed before research begins and reviewed throughout the research life cycle [111–113]. Recommended measures include independent and institutional review, containment and operational controls proportionate to the assessed risk, clear accountability and reporting requirements, responsible communication of results, and appropriate public transparency. International harmonization is also important, although implementation remains constrained by uneven resources, enforcement capacity, and the rapid development of biotechnology and AI-enabled research [13,114]. Stakeholder studies reinforce the urgency of addressing these governance gaps. In interviews with 130 participants from academia, government, industry, and policy, 76% identified AI misuse in biology as an urgent concern and 74% called for clearer governance standards; participants also emphasized the need for functional screening and more consistent oversight [97]. A separate study similarly cited concerns about engineered pathogens as a major concern and proposed specific governance measures for the convergence of synthetic biology and AI [115].

Governance of AI-enabled biology should encompass the full model life cycle, from training-data screening, predeployment evaluation, access controls, secure fine-tuning, adversarial testing, and monitoring and filtering during deployment [11,68,97,111,116,117]. Evaluations should assess capabilities such as hazardous sequence generation, virulence prediction, laboratory troubleshooting, screening evasion, agentic tool use, and uplift across different user groups. Predetermined risk thresholds should be linked to proportionate measures, including tiered access, restricted deployment, enhanced monitoring, or limits on releasing model weights. A parallel proposal extends this logic to the underlying data, where a tiered Biosecurity Data Level framework would graduate controls on biological datasets according to their AI-misuse potential [118,119]. Safeguards should also protect training data, model weights, and automated laboratory interfaces through secure development practices, authenticated data provenance, activity logging, incident reporting, and human authorization at safety-critical stages [11,68,111]. Governance must further address the digital-to-physical pathway by integrating model safeguards with customer verification, sequence screening, and oversight of commercial and benchtop synthesis. Screening should account for biological function, related orders, and hazards distributed across short fragments rather than relying only on similarity to known sequences. These responsibilities are distributed across model developers, cloud providers, synthesis companies, equipment manufacturers, laboratories, funders, regulators, and public-health agencies. Effective governance requires coordinated safeguards across actors, rather than reliance on any single institution(**Table 2**).

**Table 2. Actor-responsibility matrix for defense-in-depth governance**

| Actor | Primary obligation | Control point owned | Current gap |
|---|---|---|---|
| Model developers | Pre-deployment evaluation; secure fine-tuning; tiered release | Training data, weights, refusal (LLMs) | No standard capability thresholds; weight-locking defeatable |
| Cloud / compute providers | Enforce access tiers; monitor misuse | Deployment and inference access | Inconsistent customer verification; open-weight leakage |
| Nucleotide synthesis companies | Customer verification; function + similarity screening; cross-order aggregation | Sequence order gateway | Fragmented and AI-redesigned orders evade screening |
| Equipment / benchtop DNA synthesizer manufacturers | On-device tamper-resistant screening; secure authorization; audit | Decentralized synthesis | Benchtop devices bypass provider screening entirely |
| Laboratories / automation operators | Human authorization at safety-critical stages; sandboxing; immutable logs | Digital-to-physical execution | Weak controls on agentic and closed-loop systems |
| Funders | Conditional support on biosecurity-by-design review | Research initiation | Uneven enforcement; dual-use overlap with medicine |
| Regulators | Minimum standards; verification; incident reporting | Legal enforcement | No binding verification regime for synthesis or model access |
| Public-health agencies | Surveillance; countermeasure readiness; response | Detection and response endpoint | Under-resourced; slow validation of defensive tech |

Recent policies highlight both progress and limitations of governance for AI-enabled biological research, and the history of arms control offers precedent for the coordination this field will require. In the United States, recent policy has increased attention to synthesis screening and customer verification, while the European Union has introduced model evaluation, adversarial testing, and incident reporting obligations for providers of general purpose AI models posing systemic risk. Notably, that framework's partial exemption for openly released models mirrors the central open-weight tension in biological AI, where release decisions determine whether downstream safeguards remain enforceable at all (European Commission 2025). Internationally, the Frontier AI Safety Commitments and emerging cross-border networks of AI safety institutes support greater coordination in model evaluation, though these mechanisms remain largely voluntary. Existing mechanisms for arms control reflect similar limitations of enacting stronger safeguards. The Biological Weapons Convention prohibits biological weapons but has no verification protocol, limiting its enforcement abilities. Other regimes suggest partial models, such as the Australia Group's harmonized export controls, IAEA material accounting and inspection, and the Chemical Weapons Convention's challenge inspections.However, digital and biological artifacts are far more difficult to account for than fissile material, which is precisely why verification design, not prohibition alone, is the central unsolved problem (U.S. AI Action Plan 2025; WHO 2022).

## 6. Discussion

Biological AI capabilities are advancing rapidly across multiple dimensions, including protein design, genome engineering, and laboratory automation. Capability uplift depends critically on the interaction between model sophistication, user expertise, access to synthesis and laboratory infrastructure, and the robustness of screening and governance controls. The challenge for governance is to distinguish genuine high-consequence risks from speculative extrapolations that conflate capability with intent or operational feasibility. The near-term biosecurity priority should focus on empirically tractable threats, including AI assistance with sequence optimization and screening evasion, assembly of fragmented dangerous sequences through benchtop synthesizers, and autonomous laboratory execution of narrow but well-defined protocols. This reprioritization must be calibrated, because the same models that raise these concerns also accelerate vaccine and antimicrobial design, diagnostics, antitoxins, and pathogen surveillance; governance that suppresses capability indiscriminately forfeits defensive and biomedical value that is itself a form of biosecurity.

Two questions central to this remain open. The first is marginal risk: whether an openly released model increases capability beyond what an adversary could already obtain from published sequences and search tools. The second is foundational for governance: whether verification of biological designs and model capabilities is technically achievable, given that digital and biological artifacts are far harder to account for than the fissile material around which existing verification regimes were built.

These conclusions reflect what current systems can do, and current systems will change. As agentic, closed-loop laboratories mature, the physical-execution barrier that now separates digital assistance from real experimental capability could fall, and the autonomous experimentation results reviewed here suggest that shift is already beginning. The demonstrated-versus-speculative discipline that structures this review is therefore a tool for tracking that frontier as it moves, not a reason to assume it will move slowly.